\pdfoutput=1

\documentclass[11pt]{article}
\usepackage{overpic}

\usepackage[a-1b]{pdfx}

\usepackage{framed}
\usepackage{lmodern}
\usepackage{mdwlist}
\usepackage{siunitx}
\usepackage{latexsym}
\usepackage{colortbl}
\usepackage{xcolor}
\usepackage{nicefrac}
\usepackage{booktabs}
\usepackage{fnpct}
\usepackage{amsfonts}
\usepackage[T1]{fontenc}
\usepackage{bold-extra}
\usepackage{amsmath}
\usepackage{amssymb}
\usepackage{bm}
\usepackage{graphicx}
\usepackage{mathtools}
\usepackage{microtype}
\usepackage{multirow}
\usepackage{multicol}
\usepackage{xpatch}
\usepackage{latexsym,comment}
\usepackage[normalem]{ulem}

\newcommand*{\missingreference}{{\Huge \colorbox{red}{?reference?}}}
\newcommand*{\missingcitation}{{\Huge \colorbox{red}{?citation?}}}

\makeatletter
\xpatchcmd{\@setref}{\bfseries}{\missingreference}{}{}
\def\@citex[#1]#2{\leavevmode
    \let\@citea\@empty
    \@cite{\@for\@citeb:=#2\do
        {\@citea\def\@citea{,\penalty\@m\ }%
            \edef\@citeb{\expandafter\@firstofone\@citeb\@empty}%
            \if@filesw\immediate\write\@auxout{\string\citation{\@citeb}}\fi
            \@ifundefined{b@\@citeb}{\hbox{\reset@font\missingcitation}%
                \G@refundefinedtrue
                \@latex@warning
                {Citation `\@citeb' on page \thepage \space undefined}}%
            {\@cite@ofmt{\csname b@\@citeb\endcsname}}}}{#1}}
\makeatother

\newcommand{\gem}[1]{\mbox{\textsc{gem}}}

\newcommand{\ex}[1]{\mbox{exp}\left\{ #1\right\} }

\newcommand{\hidetext}[1]{}
\newcommand{\ignore}[1]{}
\newif\ifcomment
\commentfalse

\ifcomment
    \newcommand{\pinaforecomment}[3]{\colorbox{#1}{\parbox{.8\linewidth}{#2: #3}}}

    \newcommand{\prtodo}[1]{\pinaforecomment{lightblue}{pr}{#1}}
    \newcommand{\prtodoi}[1]{\pinaforecomment{lightblue}{pr}{#1}}
\else
    \newcommand{\pinaforecomment}[3]{}
    \newcommand{\prtodo}[1]{}
    \newcommand{\prtodoi}[1]{}
\fi

\newcommand{\smallurl}[1]{ \begin{tiny}\url{#1}\end{tiny}}

\definecolor{lightblue}{HTML}{3cc7ea}
\definecolor{CUgold}{HTML}{CFB87C}
\definecolor{grey}{rgb}{0.95,0.95,0.95}
\definecolor{ceil}{rgb}{0.57, 0.63, 0.81}
\definecolor{UMDred}{HTML}{ed1c24}
\definecolor{UMDyellow}{HTML}{ffc20e}

\usepackage[]{acl}

\usepackage{xspace}
\usepackage{xcolor}
\usepackage[utf8]{inputenc}
\usepackage{pgfplots}
\usepackage{dsfont}
\DeclareUnicodeCharacter{2212}{−}
\usepgfplotslibrary{groupplots,dateplot}
\usetikzlibrary{patterns,shapes.arrows}
\pgfplotsset{compat=newest}

\usepackage{tikzscale}
\usepackage{relsize}
\usepackage{amsmath,amssymb}
\newcommand{\probP}{\text{I\kern-0.15em P}}

\usepackage{multirow, colortbl}

\usepackage{tabularx,booktabs}
\usepackage{makecell}

\usepackage[normalem]{ulem}
\useunder{\uline}{\ul}{}

\usepackage{graphicx}
\usepackage{booktabs}

\definecolor{ablation6}{HTML}{fcefed}
\definecolor{ablation_tie}{HTML}{fce3e1}

\definecolor{ablation5}{HTML}{fcd8d4}
\definecolor{ablation4}{HTML}{FBC3BC}
\definecolor{ablation3}{HTML}{F7A399}
\definecolor{ablation2}{HTML}{F38375}
\definecolor{ablation1}{HTML}{EF6351}

\newcommand{\maxodds}{\textbf{\textsc{max odds }}}
\newcommand{\odds}{\textbf{\textsc{odds }}}

\usepackage[]{algpseudocode}
\usepackage[]{algorithm}
\usepackage{float}
\algtext*{EndFor}%
\algtext*{EndProcedure}%

\usepackage{booktabs}
\usepackage[normalem]{ulem}
\useunder{\uline}{\ul}{}

\usepackage{times}
\usepackage{latexsym}
\usepackage{adjustbox}

\usepackage[T1]{fontenc}
\usepackage{bbm}
\usepackage{amsmath}
\usepackage{amssymb}
\usepackage{booktabs}
\usepackage{tikzscale}

\usepackage{multirow, colortbl}

\usepackage{tabularx,booktabs}
\usepackage{makecell}
\usepackage{multirow}
\usepackage{scalerel,xparse}

\usepackage{cleveref}
\usepackage{microtype}
\usepackage[most]{tcolorbox}
\usepackage{gb4e}
\noautomath

\usepackage{enumitem}
\crefformat{section}{\S#2#1#3}
\crefformat{subsection}{\S#2#1#3}
\crefformat{subsubsection}{\S#2#1#3}

\definecolor{bggray}{rgb}{0.95, 0.95, 0.95}
\usepackage[%
    framemethod=tikz,
    skipbelow=\topskip,
    skipabove=\topskip
]{mdframed}
\mdfsetup{%
    leftmargin=0pt,
    rightmargin=0pt,
    backgroundcolor=bggray,
    middlelinecolor=black,
    roundcorner=3
}

\definecolor{SkyBlue}{rgb}{0.53, 0.81, 0.92}

\newtcolorbox[
  list inside=prompt,
  auto counter,
  number within=section
]{prompt}[1][]{%
  enhanced,
  float*=t,                 
  colbacktitle=black!60,
  fonttitle=\small,
  coltitle=white,
  fontupper=\footnotesize,
  boxsep=4pt,
  left=0pt, right=0pt, top=0pt, bottom=0pt,
  boxrule=1pt,
  width=\textwidth,          
  enlarge left by=0mm,
  enlarge right by=0mm,
  listing only,
  listing options={
    basicstyle=\ttfamily\footnotesize,
    breaklines=true,
    breakatwhitespace=true,
    language=json
  },
  #1,
}

\newtcolorbox[
  list inside=trace,
  auto counter,
  number within=section
]{trace}[1][]{%
  enhanced,
  float*=t,
  colback=blue!5,             
  colbacktitle=blue!60!black, 
  colframe=blue!60!black,     
  fonttitle=\small,
  coltitle=white,
  fontupper=\footnotesize,
  boxsep=4pt,
  left=0pt, right=0pt, top=0pt, bottom=0pt,
  boxrule=1pt,
  width=\textwidth,
  enlarge left by=0mm,
  enlarge right by=0mm,
  listing only,
  listing options={
    basicstyle=\ttfamily\footnotesize,
    breaklines=true,
    breakatwhitespace=true,
    language=json
  },
  #1,
}

\definecolor{UMDred}{HTML}{ed1c24}

\definecolor{yellowcolor}{HTML}{ffc20e}
\definecolor{redcolor}{HTML}{e99999}
\definecolor{orangecolor}{HTML}{f6b26b}
\definecolor{yellowcolor}{HTML}{ffd966}
\definecolor{bluecolor}{HTML}{a0c5e8}
\definecolor{purplecolor}{HTML}{d9d2e9}

\title{Filling in the Mechanisms: How do LMs Learn Filler-Gap Dependencies under Developmental Constraints?}

\author{Atrey Desai \\
  University of Maryland\\
  \texttt{adesai10@umd.edu} \\\And
  Sathvik Nair \\
  University of Maryland \\
  \texttt{sathvik@umd.edu}}

\begin{document}
\maketitle

\begin{abstract} {
For humans, filler-gap dependencies require a shared representation across different syntactic constructions. Although causal analyses suggest this may also be true for LLMs \citep{boguraev2025causal}, it is still unclear if such a representation also exists for language models trained on developmentally feasible quantities of data.
We applied Distributed Alignment Search (DAS, \citet{geiger2024finding}) to LMs trained on varying amounts of data from the BabyLM challenge \citep{warstadt2023findings}, to evaluate whether representations of filler-gap dependencies transfer between wh-questions and topicalization, which greatly vary in terms of their input frequency.
Our results suggest shared, yet item-sensitive mechanisms may develop with limited training data. More importantly, LMs still require far more data than humans to learn comparable generalizations, highlighting the need for language-specific biases in models of language acquisition.\footnote{Our code and data are available at: \url{https://github.com/atreydesai/developmental-filler-gap}}

}
\end{abstract}

\section{Introduction} \label{section:intro}
A major question in language acquisition asks how learners can make generalizations about infinite utterances on the basis of finite input. 
Many cognitive models have proposed and revised many claims about the linguistic representations humans could use to solve this learning problem, which often apply abstract, language-specific rules to learn from specific pieces of data \citep{yang2004universal,perkins2022power,pearl2023computational}.
Language models (LMs), on the other hand, lack these biases, yet still posit some important syntactic generalizations \citep{futrell2019neural,warstadt2020blimp,linzen2021syntactic,wilcox2024using}. 
These successes have led some researchers to question the need for language-specific representations altogether \citep{piantadosi2023modern,futrell2025linguistics}.

One particular case that evaluates this debate involves \textit{filler-gap dependencies}.
Filler-gap dependencies are a type of syntactic relation formed when a constituent (or the filler) is displaced from its canonical position (the gap) and interpreted in another position. These gaps can exist across various constructions including, but not limited to, wh-questions (\textit{``What did the student make \_?''}), relative clauses (\textit{``the robot that the student made \_''}), and topicalization (\textit{``This robot, the student made \_''}).

Filler-gap dependencies are a useful test case for evaluating the representations necessary for language acquisition, as they not only require learners to recognize hierarchical structure across items in a sentence, but also notice an empty syntactic position that is not overtly realized.
Many linguistic theories claim that despite superficial differences, these constructions may share an abstract underlying mechanism \citep{chomsky_wh-movement_1977,culicover1977formal,gazdar_phrase_1982,kaplan_lexical-functional_1982,postal_three_1999}. 
Evidence from real-time processing supports these claims; adults exhibit similar behavioral patterns across multiple filler-gap constructions \citep[among others]{crain_rules_1985,traxler1996plausibility,sprouse2016experimental,kush2021sentence}.
Determining how a generalized representation can be learned from linguistic data is thus a major area of investigation.

LMs are relevant to addressing this question since they are an example of a ``domain-general, weakly biased'' learner \citep{wilcox2024using}. Their successes with some constraints on filler-gap dependencies have led to claims that the dependency \textit{is} learnable without language-specific knowledge \citep{wilcox2018rnn,wilcox2024using}.
These results have since been challenged, highlighting that LMs may not represent filler-gap dependencies in a human-like manner \citep{bhattacharya-van-schijndel-2020-filler,lan2024large,howitt2024generalizations,chang2025mind}. Although analyses of LM probabilities show mixed results, causal interventions on LMs' internal states could tell a more definitive story as to whether a shared representation of filler-gap dependencies persists across constructions. \citet{boguraev2025causal} do apply causal interpretability methods \citep{geiger2024finding,mueller2025quest}, and identify shared underlying structure across various types of filler-gap dependencies.
However, even if \citet{boguraev2025causal} show a shared representation of filler-gap dependencies may be learnable in principle, their results do not address whether inductive biases are needed for humans, as they evaluated LMs trained on data that matches neither the content nor the quality of data available to human learners \citep{wilcox2025bigger}. \citet{chang2025mind} do investigate LMs trained on human-scale data, but rely on probability measures for one type of filler-gap dependency.

In our study, we run causal interventions on LMs \citep{boguraev2025causal} trained on data from the BabyLM challenge \citep{warstadt2023findings}, which reflects material that English-speaking children could be exposed to, up to 12 years of age. 
Our experiments show that LMs may posit a shared representation of filler-gap dependencies across high and low frequency constructions,  but rely on far more input than human learners \citep{perkins2021eighteen}.
This representation is also far less general than the ones proposed by linguists, showing systematic variability across items and constructions.
Overall, our results suggest that human learners need a combination of domain-general statistical learning mechanisms along with language-specific inductive biases \citep{yang2004universal,portelance2024roles} to learn relevant linguistic generalizations. Developmentally aware evaluation should also include timing of human acquisition alongside the amount and type of training data. Children demonstrate sensitivity to core syntactic structure quite early, such as filler-gap representations by 18 months~\cite{perkins2021eighteen}. A model is only developmentally plausible if comparable generalizations emerge on a similarly early developmental timescale.
\section{Background} \label{section:background}
Filler-gap dependencies are shared across syntactic constructions that show different surface forms, even if they may serve different semantic and discourse functions \citep{schutze_challenges_2015}.
Looking at the following sentences, (\ref{ex:wh}) is a \textit{wh-question}, while (\ref{ex:topic}) is an example of \textit{topicalization}.

\begin{exe}
\ex \textit{Who} did the teacher like \_ ?
\label{ex:wh}
\ex Did the teacher like?*
\label{ex:wh*}
\ex \textit{The author}, the teacher liked \_ .
\label{ex:topic}
\ex The teacher liked.*
\label{ex:topic*}
\end{exe}

In (\ref{ex:wh}) \textit{who} is the object of \textit{like}, and in (\ref{ex:topic}), \textit{The author} is the object of \textit{liked}. These constituents, which are \textbf{fillers}, are fronted to form a dependency with the \textbf{gaps}, which are unpronounced but marked with \_ for readability. 
Learning the generalization also involves recognizing where the dependency may \textit{not} be valid, such as (\ref{ex:wh*}) and (\ref{ex:topic*}), which show that (\ref{ex:wh}) and (\ref{ex:topic}) are respectively ungrammatical without the fillers.
Syntactic configurations called \textit{islands} make extracting a filler ungrammatical \citep{chomsky_wh-movement_1977}. \footnote{Refer to \citet{wilcox2024using}, \citet{howitt2024generalizations}, and \citet{chang2025mind} for further discussion relating to language models and island constraints.}
This has made recognizing filler-gap licensing a particularly relevant test case when evaluating syntactic structure in language models.

\subsection{LM Surprisal and Filler-Gap Dependencies}

Many studies use LMs to compute the \textit{surprisal}, or negative log probability, of a word given its context. Surprisal quantifies the effect of processing difficulty \citep{levy2008expectation}, and evaluating LMs' surprisals at particular points in a sentence effectively identifies which parts are expected to be more difficult to process \citep{futrell2019neural}.\footnote{However, see \citet{huang2024large} for evidence that LM surprisal cannot reflect the quantitative effects of processing difficulty for particular types of syntactically complex sentences.} 
Work evaluating LMs on syntactic structure has often relied on comparing two minimal pairs of sentences, such as (\ref{ex:wh}) vs. (\ref{ex:wh*}) and (\ref{ex:topic}) vs. (\ref{ex:topic*}), where the ungrammatical version should typically have a higher surprisal than the grammatical version \citep{marvin2018targeted,warstadt2020blimp,gauthier2020syntaxgym}.

In English sentences with embedded wh-movement, LSTM language models have shown positive results in recognizing the presence and absence of fillers and gaps \citep{wilcox2018rnn}. These results have been extended to Transformer models and various island constraints \citep{wilcox2024using}. \footnote{Results are more mixed in Norwegian \citep{kobzeva_neural_2023} and Dutch \citep{suijkerbuijk2023learnability}, which have different filler-gap structures from English.}

\citet{ozaki2022well} evaluated LSTMs across a range of other filler-gap constructions, including topicalization, and found that model performance for each construction is highly correlated with its frequency.
They do not present evidence whether this generalization is shared \textit{across} constructions. 

To this end, other approaches involve providing LMs with additional training examples. 
Simulated priming studies on wh-movement \citep{bhattacharya-van-schijndel-2020-filler,prasad2019using} have shown some evidence for a shared representation of filler-gap dependency, but not constraints on the dependency.
More recent studies have retrained LMs by augmenting their training data with positive examples of a dependency.
\citet{lan2024large} show that augmentation improves LMs' performance on complicated filler-gap constructions (parasitic gaps and across-the-board movement).
Extending this approach across constructions, \citet{howitt2024generalizations} adopted their methodology and found that augmenting LSTMs' training data with instances of one construction (clefting and topicalization) failed to improve performance on detecting filler-gap licensing and islands for other constructions (wh-movement, clefting, topicalization, and tough-movement).

Although surprisal can effectively show whether LMs are able to assign probabilities appropriately for cases of filler-gap licensing, it may not necessarily directly reflect LMs' internal representations. Retraining LMs can help address whether they posit a shared representation for filler-gap dependencies, but results have been mixed.

\subsection{Causal Interventions}
In order to evaluate whether LM representations can indeed encode linguistic features of interest, several studies have adopted methods from mechanistic interpretability, using \textit{causal interventions}. Broadly speaking, these approaches manipulate an LM's internal representations of individual items to find aspects of its representational space that are causally responsible for a desired feature \citep[among others, see \citet{mueller2025quest} for a review]{wanginterpretability,lasri2022probing,hao2023verb,kryvosheieva2025different}. 
\citet{arora2024causalgym} evaluated causal intervention methods on models from the Pythia series \citep{biderman2023pythia}, run on a large-scale suite of syntactic constructions \citep{gauthier2020syntaxgym}, including wh-movement. 
\citet{arora2024causalgym}'s best-performing method, Distributed Alignment Search (DAS) \citep{geiger2024finding}, learns a rotation of the representation space to identify a direction; when intervened upon, this vector maximizes the probability of a counterfactual output label. We provide further description of this method in \ref{sec:das_method}. \citet{boguraev2025causal} applied DAS to Pythia-6.9B, finding evidence for a shared representation across seven types of filler-gap constructions.
In their analysis, they identify an underlying ``transfer network'' that evaluates whether the representation from one filler-gap construction can successfully be implanted in another.
More frequent constructions, such as wh-questions, act as ``source’’ nodes in the network towards infrequent ``sink’’ nodes such as topicalization. Despite unequal contributions to the shared representation, evidence for such a representation did exist. 
They also identify a ``lexical boost''\footnote{This term is borrowed from psycholinguistic studies using syntactic priming, where human processing gets facilitated by similar sentence structures \citep{traxler2014syntactic}.} effect based on whether the examples involved in the intervention share the same level of animacy, that is; stronger causal effects when both sentences in the intervention describe animate or inanimate subjects. 

\subsection{Developmentally Plausible Data}
Although \citet{boguraev2025causal} shows evidence for an LM positing a shared mechanism for filler-gap dependencies, they relied on a Transformer model with billions of parameters, trained on a Web-scale corpus \citep{gao2020pile}.
These results make the case that filler-gap dependencies are learnable with a domain-general mechanism in principle, but have less to say about questions of \textit{human} language acquisition \citep{wilcox2025bigger}.
Children are able to understand constraints on filler-gap dependencies around ages 3-5 \citep[among others]{de1995relative,friedmann2009relativized}, while more recent evidence suggests they begin to demonstrate this sensitivity as early as 18 months \citep{gagliardi2016discontinuous,atkinson2018developing,perkins2021eighteen}. \footnote{Most of this work relies on wh-questions and relative clauses to argue children rely on a shared representation of filler-gap constructions. However, a recent cross-linguistic corpus study finds evidence for topicalization in children's utterances around the age of 2 \citep{bosch2025another}.}

The BabyLM Challenge \citep{warstadt2023findings} trains models on approximately 100 million tokens, the estimated linguistic input of a child around 12 years of age \citep{gilkerson2017mapping}. 
The data and models from BabyLM allow researchers to answer questions about human language acquisition that would be difficult to test experimentally, while providing learners with training data mimicking children's input.
A prior evaluation of BabyLM models' performance on filler-gap licensing and island constraint violations found partial but incomplete acquisition, with performance varying substantially across island types \citep{chang2025mind}. 
However, their findings are limited to wh-movement, which frequently occurs in children's input \citep{furrow1979mothers}, but models should also be evaluated across other constructions that appear less frequently to see if they posit a shared representation.

\begin{figure*}[t]
\centering
\includegraphics[width=\textwidth]{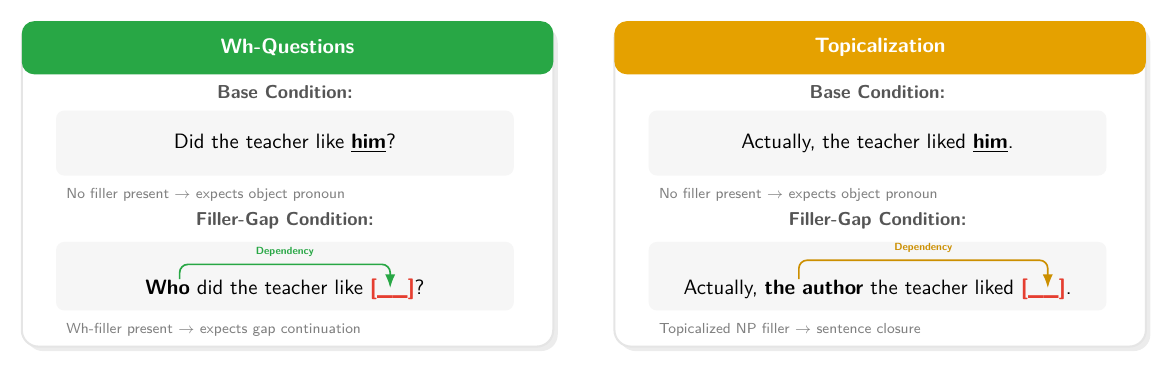}
\caption{The diagram contrasts Wh-Questions (left, green) and Topicalization structures (right, orange). }
\label{fig:experiments}
\end{figure*}
\section{Research Questions and Hypotheses}
Our study seeks to evaluate whether a causal cross-construction representation for filler-gap dependencies is learnable under a domain-general mechanism, provided that a learner has access to child-like amounts of input data.
To this end, we aim to replicate findings from \citet{boguraev2025causal} on a model trained on the corpus used in the BabyLM challenge, in service of extending results from linguistically informed mechanistic interpretability studies to questions about language development.
We evaluate the model on both a high-frequency construction: wh-questions, compared to a low-frequency construction, topicalization.
We propose the following three research questions:
\begin{itemize}
    \item \textbf{RQ1:} Can a language model with relatively few parameters and human-like training data still posit a causal representation of filler-gap dependencies? If so, when does this effect emerge? 
    \item \textbf{RQ2:} Are representations localized within examples of the same construction?
    \item \textbf{RQ3:} Do representations transfer from low-frequency to high-frequency constructions? 
\end{itemize}
We create three hypotheses based on these questions. 
First, the version of the model trained on all 100 million tokens in the BabyLM training corpus \textit{should} learn an abstract filler-gap mechanism that is detectable and transferable by DAS from one construction to another, based on some of the positive results from \citet{chang2025mind}. 
However, due to the lack of language-specific inductive biases, we do not predict a strong causal effect before 10 million tokens. This is because the BabyLM checkpoint for 10 million tokens mimics the linguistic input of children from ages 2-5 years \citep{warstadt2023findings}, while English-speaking children are able to recognize filler-gap dependencies around 18 months \citep{perkins2021eighteen} (\textbf{H1: Misaligned Emergence Hypothesis}). 
Second, since LMs likely learn the filler-gap dependency in a piecemeal fashion \citep{ozaki2022well,howitt2024generalizations}, we expect stronger causal effects for within-construction interventions compared to cross-construction interventions. We also hypothesize that transfer improves when both constructions share the same level of animacy, to replicate \citep{boguraev2025causal}'s findings for lexical boost effects.
 (\textbf{H2: Construction-Specificity Hypothesis}).
Third, given the greater prevalence of wh-questions in language corpora, as opposed to topicalization, and existing work showing learning correlates with input frequency \citep{ozaki2022well,boguraev2025causal}, we expect an asymmetrical and one-way transfer from high-frequency wh-questions to low-frequency Topicalization (\textbf{H3: Frequency Modulation Hypothesis}).

\section{Methods} \label{section:methods}
\subsection{Model}
We use the BabyLM-100M model~\cite{warstadt2023findings}. This model uses the GPT-2-small architecture \citep{radford2019language} trained on the BabyLM Strict-100M corpus, consisting of a set of approximately 100 million words as training data designed to mimic the total linguistic input received by an English-speaking child until early adolescence (around 12 years of age) \citep{gilkerson2017mapping,warstadt2023findings}.

The corpus consists of relevant data from the British National Corpus, CHILDES language acquisition database, Switchboard Dialog Act Corpus, subtitles from children’s TV shows, and simplified Wikipedia articles~\cite{charpentier2025babylmturns3papers}.

Multiple checkpoints for the model across the training process were also released. We evaluate several checkpoints where the model received increasing amounts of input (1M--100M tokens): 10 checkpoints from 0--10M tokens and 9 checkpoints between 10M--100M tokens. Additionally, nine additional checkpoints (100M--1000M) are used in extended analysis in the appendix.

\subsubsection{Constructions}
We use two filler-gap constructions that have different levels of frequency: matrix wh-questions (high frequency) and topicalization (low frequency), based on \citet{ozaki2022well}. This combination allows us to systematically evaluate whether knowledge of high-frequency constructions can be transferred to less frequent ones that may barely be present in children's input.

\textbf{Wh-Questions (High Frequency).} We use single-clause matrix wh-questions, such as \textit{``What did the doctor do \_\_\_?''} and \textit{``What did the student read \_\_\_?’’} Wh-questions are very common in natural language and exist in child-directed speech \citep{furrow1979mothers}. Prior work demonstrates that neural LMs readily acquire sensitivity to wh-dependencies in English \citep{wilcox2024using}. This makes Wh-questions a plausible construction for a high-frequency ``source'' in transfer experiments.

\textbf{Topicalization (Low Frequency).} We employ fronted object topicalization with an optional discourse marker, such as \textit{``The student, the teacher liked \_\_\_''} or \textit{``Actually, the book the author read \_\_\_''} Topicalization is extremely uncommon in natural language and almost absent from child-directed speech \citep{roland2007frequency}. Related works find that LMs fail to display the correct behavior for topicalization \citep{ozaki2022well}, even when augmented with examples in the training data ~\cite{howitt2024generalizations}.\footnote{\citet{howitt2024generalizations} show that including the discourse marker leads to slight qualitative improvements, but the generalization is only learned in one direction, when the filler is present.} In addition, when analyzing transfer of the filler-gap mechanism across constructions, \citet{boguraev2025causal} found topicalization had a low ``out-degree'' in their transfer network: despite receiving other constructions, topicalization rarely transfers to them. This makes topicalization a strong candidate for a low-frequency ``sink'' construction for our experiments.

We created minimal pairs following the original template and methodology of \citet{boguraev2025causal}. Each pair compares a base sentence with no filler-gap dependency with a filler-gap sentence containing a matrix wh-question or a topicalization setup. The model’s expected next-token prediction differs based on the two predicted gaps (marked with \_):

\subsection{Materials}
Replicating the experimental structure of \citet{boguraev2025causal}, we test sentences with both animate and inanimate fillers. Animate fillers use \textit{who} (wh-questions) or NPs with perceived life or agency, such as \textit{the author} (topicalization). In contrast, inanimate fillers use {what} (wh-questions) or nonliving and nonsentient NPs such as \textit{the book}. This creates four dataset template variants: \textit{wh\_animate}, \textit{wh\_inanimate}, \textit{topic\_animate}, and \textit{topic\_inanimate}.

Different combinations of the following lexical items are used in respective templates:
\textbf{Subject NPs} (50 animate nouns: \textit{teacher, doctor, manager}, etc.), 
\textbf{Verbs} (30 transitive verbs: \textit{like, admire, follow}, etc.), 
\textbf{Auxiliaries} for wh-questions (7 verbs: \textit{did, will, could}, etc.), and
\textbf{Licensing adverbs} for topicalization (25 adverbs: \textit{Actually, Frankly, Surprisingly}, etc.).
We verified all materials consist of one token for the model and occur within the BabyLM corpus.\footnote{See \citet{nair2023words}, \citet{giulianelli_proper_2024}, and \citet{oh2025impact} for discussion of tokenization and psycholinguistic applications.}

This yielded approximately 21,000 unique sentence pairs for wh-questions and 1,875,000 for topicalization per animacy condition. Topicalization randomly selects both the sentence-initial adverb and the topicalized filler phrase, increasing the amount of unique sentences. However, both pools remain substantially larger than 2000 pairs sampled for DAS training, maintaining sentence diversity for generalization. 

\subsection{Distributed Alignment Search (DAS)}
\begin{figure}[t]
\centering
\includegraphics[width=0.48\textwidth]{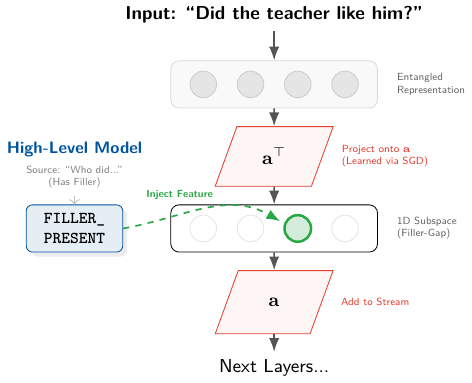}
\caption{To create a DAS vector, we learn a direction $\mathbf{a}$ to align neural representations with the binary variable \texttt{FILLER\_PRESENT}. We intervene by projecting the difference between the source and base representations onto $\mathbf{a}$ and injecting it into the base sentence.}
\label{fig:das}
\end{figure}

\label{sec:das_method}
Distributed Alignment Search (DAS) is a causal intervention method to test if a high-level concept aligns with the internal weights of a language model~\cite{geiger2024finding}. We can define a minimum causal model for a filler-gap dependency using a binary variable: $\textsc{Filler\_Present} \in \{0, 1\}$. This variable causally influences gap expectations and the model's next-token predictions. 

Given a \textit{source} sentence with a filler (\ref{fig:experiments}) and a \textit{base} sentence without a filler
(\ref{fig:experiments}), we can intervene based on the internal representations of the base sentence by implanting the learned filler-gap DAS feature from the source, as seen in figure \ref{fig:das}. A successful implementation should shift the prediction of the model from the base label (\textit{him}) toward the source label (\textit{?}). This would suggest that the filler-gap dependency is encoded at the intervention site~\cite{wu-etal-2024-pyvene}.

\subsubsection{Training}
Following prior work, we use a 1-dimensional variant of DAS vector~\cite{geiger2024finding,arora2024causalgym,boguraev2025causal}. Given an embedding space with dimensionality $d$, for each layer $\ell$ and token position $p$, we learn a \textit{direction vector} $\mathbf{a}_{\ell,p} \in \mathbb{R}^d$ that defines a one-dimensional subspace in which the filler-gap feature is encoded, between the model's representations of the base construction at $\ell$ and $p$ ($\mathbf{h_{\text{base},\ell, p}} \in \mathbb{R}^d$) and the source construction ($\mathbf{h_{\text{source},\ell, p}} \in \mathbb{R}^d$). Once the vector $\mathbf{a}$ is learned, the intervention projects the difference between the source and base representations onto $\mathbf{a}$ and adds it to the base representation:

\begin{equation}
    \tilde{\mathbf{h}} = \mathbf{h}_{\text{base}} +\mathbf{a}\mathbf{a}^\top (\mathbf{h}_{\text{source}} - \mathbf{h}_{\text{base}})
\end{equation}

This intervention preserves the orthogonal dimensions of the base representation and only modifies the value along the learned feature direction $\mathbf{a}$. The direction is optimized to minimize cross-entropy loss between the counterfactual predictions of the model after intervention and the source sentence labels. 

We use a batch size of 25, 80 training steps per layer-position combination, and a learning rate of 5$\times$10$^{-3}$ to train each DAS vector. We provide further information on hyperparameter selection in Appendix~\ref{section:appendix}. Due to the comparatively low number of layers in BabyLM's architecture, we test all 12 layers across 6 token positions aligned to template slots: prefix (position 0),  filler (position 1), auxiliary/complementizer (position 2), article (position 3),  subject NP (position 4), and verb (position 5).

\begin{figure*}[t]
\centering
\includegraphics[width=\textwidth]{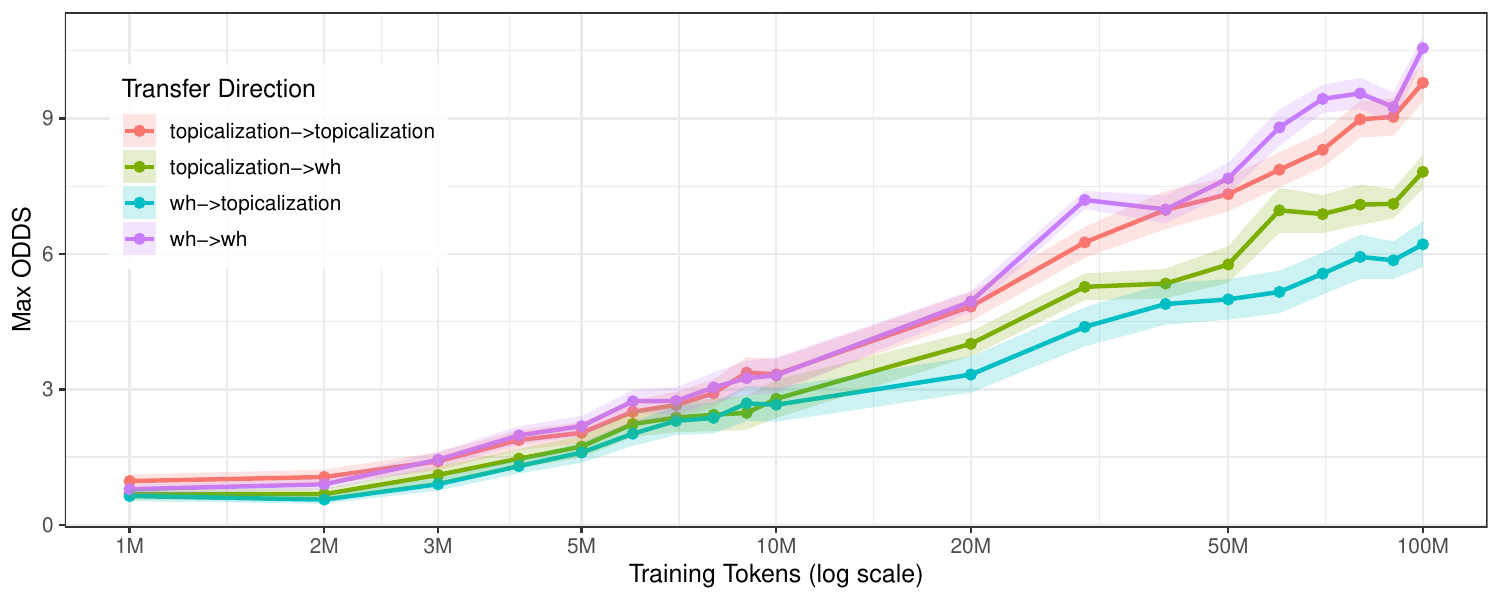}
\caption{Developmental trajectory of all filler-gap mechanisms across training. Error bands show $\pm$1 SE across a minimum of 6 seeds. All four conditions show a monotonic increase with training tokens.}
\label{fig:emergence}
\end{figure*}

\subsubsection{Evaluation Metrics}
We primarily use the \odds metric to quantify the magnitude of the causal effect through measuring how much the intervention shifts log-probabilities toward the counterfactual outcome. This is given by the formula:

\begin{equation}
\begin{aligned}
\text{\odds} &= \log \frac{P(\text{base} \mid \text{clean})}{P(\text{source} \mid \text{clean})} \\
            &+ \log \frac{P(\text{source} \mid \text{int})}{P(\text{base} \mid \text{int})}
\end{aligned}
\label{eq:odds}
\end{equation}

In Equation~\eqref{eq:odds}, `clean' refers to a standard forward pass without an intervention, while `int' refers to a forward pass where the DAS intervention is applied.

A positive \odds value suggests the intervention successfully shifts predictions toward the source; in addition, the higher the \odds values, the stronger the causal effects. 

Utilizing empirical findings from \citet{arora2024causalgym} on the Pythia model family, we establish the following qualitative thresholds: values near 0 show little to no causal effect (comparable to random baselines), values in the 3--6 range demonstrate emerging to moderate causal structure (as seen in smaller models such as Pythia-14M), and values greater than 8 indicate strong causal mechanisms (as seen in larger models such as Pythia-6.9B).

We primarily report \maxodds, or the maximum \odds value across all layers at a given position. The goal of DAS is to localize the feature to specific layers, so the maximum represents the layer-position combination with the most effective causal effect.

\subsubsection{Experiments}
We run two types of experiments: \textit{localization} and \textit{transfer}, to evaluate generalizations within and across construction types.
\begin{enumerate}
    \item \textbf{Wh $\rightarrow$ Wh (within-construction localization)}: Train DAS on wh-questions, test on held-out wh-questions
    \item \textbf{Topic $\rightarrow$ Topic (within-construction localization)}: Train DAS on topicalization, test on held-out topicalization
    \item \textbf{Wh $\rightarrow$ Topic (forward transfer)}: Train DAS on wh-questions, test on topicalization
    \item \textbf{Topic $\rightarrow$ Wh (backward transfer)}: Train DAS on topicalization, test on wh-questions
\end{enumerate}
The localization experiments (Wh$\rightarrow$Wh, Topic$\rightarrow$Topic) quantify the extent to which the DAS can identify filler-gap representations within each construction type. Likewise, cross-construction transfers (Wh$\rightarrow$Topic, Topic$\rightarrow$Wh) test if representations of a given construction are generalizable across the filler-gap constructions. If the transfer is symmetric, \odds retention should be similar in both directions. If frequency modulates transfer, we predict asymmetrical transfer towards the high-frequency (Wh $\rightarrow$ Topic) direction.

\subsection{Statistical Analysis}
We fit a linear model predicting \maxodds from number of training tokens, transfer direction, and animacy:

\begin{equation}
    y = \beta_0 + \beta_t \mathbf{x}_{\text{tokens}} + \beta_d \mathbf{x}_{\text{dir}} + \beta_a \mathbf{x}_{\text{anim}} + \epsilon
\end{equation}

Where $y$ is \maxodds, and $\mathbf{x}$ represents the fixed effects for token count, transfer direction, and animacy. Post-hoc contrasts used estimated marginal means with Holm-Bonferroni correction for 171 pairwise comparisons~\cite{Lenth2017}. Cohen's $d$ is used to measure effect sizes using the residual standard deviation. 

\section{Results} \label{section:results}
We present the results from 19 BabyLM checkpoints in the developmentally plausible range (1M--100M tokens) across multiple constructions (wh-questions, topicalization) and animacy conditions (animate, inanimate). All experiments were repeated on a minimum of 6 seeds to establish tighter bounds of confidence. The linear model achieved  $R^2 = 0.53$, $F(22, 4537) = 235$, $p < .001$.

\subsection{Developmental Trajectory}
Figure~\ref{fig:emergence} shows \maxodds across training on all four transfer directions. Filler-gap localization significantly increased with training duration ($F(18, 4537) = 264.3$, $p < .001$), from near zero at 1M tokens (\maxodds $\approx$ 0.8) to a robust effect by 100M (\maxodds $\approx$ 10.6 for Wh$\rightarrow$Wh).
Qualitatively, relatively stronger causal effects (\maxodds > 8) began emerging after around 50M tokens, while the effects are weaker (\maxodds $\approx$ 3) around 10M tokens. 

\subsection{Within-Construction Localization}
Both constructions displayed successful within-construction localization, increasing with training tokens. At 100M tokens, Wh$\rightarrow$Wh achieved \maxodds = 10.56 (SD = 2.11) and Topic$\rightarrow$Topic achieved \maxodds = 9.79 (SD = 3.24). Compared to the results of the transfer experiments, within-construction localization effects consistently exceeded cross-construction transfer effects (mean difference = 1.33 \maxodds, $t$ = 17.4, $p < .001$,  $d$ = 0.52), showing that high performance on one dependency type does not completely generalize to others.

\subsection{Cross-Construction Transfer}
Unlike the frequency-based prediction in Hypothesis 3, Topic$\rightarrow$Wh 
transfer \textit{exceeded} Wh$\rightarrow$Topic transfer throughout 
training (difference = 0.57 \maxodds, $t$ = -5.29, $p < .001$, $d$ = -0.22). At the 100M checkpoint, Topic$\rightarrow$Wh achieved \maxodds = 7.82 
versus Wh$\rightarrow$Topic's \maxodds = 6.21. 

This transfer asymmetry could be due to wh-questions developing a more construction-specific representation that transfers less effectively across other types of filler-gap constructions. Conversely, a more general mechanism may be reflected in topicalization when it is learned, due to little to no presence in the input. In an exploratory analysis, we also evaluate later checkpoints for the model as it was trained on more data, and find that the trend does not persist. Instead, two types of cross-construction transfer have similar causal effects on models with more input (Appendix \ref{sec:extended}).

\subsection{Animacy Effects}
\label{sec:animacy_effects}

We find further evidence of a ``lexical boost'' effect predicted by \citet{boguraev2025causal}. DAS transfer was significantly stronger when animacy of training and evaluation matched relative to animacy-mismatched conditions (difference = 0.67 \maxodds, $t$ = 4.86, $p < .001$, $d$ = 0.28). 

\begin{table}[t]
\centering
\small
\setlength{\tabcolsep}{3.5pt} 
\begin{tabular}{lccccc}
\toprule
\textbf{Contrast} & \textbf{Est.} & \textbf{SE} & $t$ & $p$ & $d$ \\
\midrule
Within--Across & 1.33 & 0.08 & 17.4 & <.001 & 0.52 \\
Wh→Topic--Topic→Wh & -0.57 & 0.11 & -5.3 & <.001 & -0.22 \\
Animate--Inanimate & 0.68 & 0.08 & 8.6 & <.001 & 0.26 \\
\bottomrule
\end{tabular}
\caption{Post-hoc comparisons from linear model. Negative asymmetry results
suggest Topic$\rightarrow$Wh outperformed Wh$\rightarrow$Topic.}
\label{tab:contrasts}
\end{table}

\section{Discussion and Conclusions} \label{section:conclusion}
This study applies causal interventions to determine how a language model learns filler-gap dependencies when provided with developmentally realistic amounts of training data from the BabyLM corpus, determining if findings from larger LMs \citep{boguraev2025causal} apply in this setting.
Using DAS, we evaluated BabyLM-100M's generalizations in four experimental conditions: localization within constructions and transfer across constructions, for two types of filler-gap dependencies, wh-questions and topicalization. 
Our results show that the model learns a shared representation for filler-gap dependencies, but still requires far more data than children would, and is still highly sensitive to variation across constructions.

Regarding \textbf{RQ1}, which asks both whether and when LMs learn a causal representation, we find evidence in favor of our misaligned emergence hypothesis.
Although the BabyLM-100M model showed strong causal effects when trained on the full corpus, they failed to emerge when the model received human-like quantities of training data.
The full corpus was comparable to the input available to English-speaking adolescents (up to around 12 years) \citep{warstadt2023findings}, while children's sensitivity to filler-gap dependencies emerges prior to two years of age, and robust knowledge develops between the ages of 3 and 5. 
In their description of the BabyLM corpus, \citet{warstadt2023findings} report that the 10M checkpoint corresponds to children's linguistic knowledge between the ages of 2 and 5. 
If the model \textit{was} learning with a human-like mechanism, we would expect moderate causal effects prior to this checkpoint, and strong causal effects at 10 million tokens, yet we only identify weak effects, if any.

\textbf{RQ2} asks whether the learned representations are specific to particular constructions. We found evidence for this construction-specificity hypothesis because the localization experiments consistently performed better than the transfer experiments. That is, generalization within examples of the same construction was far more effective than transferring representations across constructions. We also replicated the lexical boost effects when animacy gets matched during the intervention, suggesting this feature may transfer across items. Regarding the direction of transfer, as discussed in \textbf{RQ3}, we found improved performance generalizing from topicalization to wh-questions, which was the opposite of our predictions in the frequency modulation hypothesis. 
Future work can determine if this happens because LMs' may learn more item-sensitive representations of filler-gap constructions early during training, only generalizing these representations after receiving far more input than humans.

Overall, instead of positing a single, general representation of filler-gap constructions, LMs learn item and construction-specific representations.
Future work should extend DAS to evaluate learning the filler-gap dependency in both directions and sensitivity to island constraints, across more diverse construction types.

When modeling human language acquisition, however, our results show LMs trained solely on next-word prediction are not sufficient to learn appropriate syntactic generalizations with human-like input.
Instead, we emphasize the need to model learning with explicit inductive biases over structured hypothesis spaces, in the spirit of \citet{perkins2025mind,portelance2025reframing}.
LMs can still play a role in this enterprise, through reflecting inductive biases architecturally \citep{murty2023pushdown} or specifying possible hypothesis spaces \citep{misra2024generating,portelance2024roles}. 
Our work joins a conversation \citep{yang2026unified,zhou2026exactly} about the needs to further constrain models of language acquisition to better match human behavior by emphasizing how even developmentally constrained LMs require superhuman amounts of input to make correct linguistic generalizations.

\section*{Limitations} \label{section:limitations}
This study only focused on English, limiting the generalizability of these results to other languages where filler-gap dependencies behave differently under LMs \citep{kobzeva_neural_2023,suijkerbuijk2023learnability}.
Additionally, the training corpus is based on text input alone, while children learn from spoken data, multimodal environments, and social interaction \citep{meylan2023adults,vong2024grounded}.
Since this study focused on extending \citet{boguraev2025causal}'s results to a BabyLM-scale model, we did not evaluate whether it could recognize the absence of filler-gap dependencies, and for island constraints, which have been used in surprisal-based studies \citep{ozaki2022well,wilcox2024using,howitt2024generalizations,chang2025mind}.
More complex materials would also be useful to ensure results are not associated with confounding factors like punctuation, since we extract representations from periods and question marks.
Lastly, although DAS was the best performing method from \citet{arora2024causalgym}, measures like Boundless DAS operate over subspaces instead of single dimensions \citep{wu2023interpretability,geiger2024finding}, and could have yielded stronger causal effects. 

\section*{Ethical Considerations}
This work used publicly available data and models, which are described further in the original publications.
We do not foresee any risks associated with this work, as we used the data for their intended purpose to study human language acquisition.
Generative AI (GenAI) was used in this project.
We used Antigravity\footnote{https://antigravity.google/} to design plots and refactor code, and Claude Opus 4.5 to refine paper writing for brevity.
We never use GenAI for writing text from scratch in this paper.
We take complete responsibility for any GenAI errors.
By discussing GenAI usage here, we aim to encourage other researchers to do the same.

\section*{Acknowledgements}
We would like to thank Alba Jorquera, Katherine Howitt, Jeffrey Lidz, Philip Resnik, Omar Agha, Samer Nour Eddine, Kartik Ravisankar, Navita Goyal, and Rupak Sarkar from UMD's Linguistics Department, Computational Cognitive Science group \& CLIP lab, and audiences at the Texas Linguistics Society conference for helpful discussions and feedback on this work.
This material is based on work supported by the NSF GRFP (No.~DGE 2236417) to Sathvik Nair. Any opinions, findings, and conclusions or recommendations expressed in this material are those of the authors and do not necessarily reflect the views of the National Science Foundation.

\bibliography{custom}

@inproceedings{wilcox2018rnn,
    title = "What do {RNN} Language Models Learn about Filler{--}Gap Dependencies?",
    author = "Wilcox, Ethan  and
      Levy, Roger  and
      Morita, Takashi  and
      Futrell, Richard",
    editor = "Linzen, Tal  and
      Chrupa{\l}a, Grzegorz  and
      Alishahi, Afra",
    booktitle = "Proceedings of the 2018 {EMNLP} Workshop {B}lackbox{NLP}: Analyzing and Interpreting Neural Networks for {NLP}",
    month = nov,
    year = "2018",
    address = "Brussels, Belgium",
    publisher = "Association for Computational Linguistics",
    url = "https://aclanthology.org/W18-5423/",
    doi = "10.18653/v1/W18-5423",
    pages = "211--221"
}

@inproceedings{howitt2024generalizations,
    title = "Generalizations across filler-gap dependencies in neural language models",
    author = "Howitt, Katherine  and
      Nair, Sathvik  and
      Dods, Allison  and
      Hopkins, Robert Melvin",
    editor = "Barak, Libby  and
      Alikhani, Malihe",
    booktitle = "Proceedings of the 28th Conference on Computational Natural Language Learning",
    month = nov,
    year = "2024",
    address = "Miami, FL, USA",
    publisher = "Association for Computational Linguistics",
    url = "https://aclanthology.org/2024.conll-1.21/",
    doi = "10.18653/v1/2024.conll-1.21",
    pages = "269--279"
}

@inproceedings{biderman2023pythia,
  title={Pythia: A suite for analyzing large language models across training and scaling},
  author={Biderman, Stella and Schoelkopf, Hailey and Anthony, Quentin Gregory and Bradley, Herbie and O’Brien, Kyle and Hallahan, Eric and Khan, Mohammad Aflah and Purohit, Shivanshu and Prashanth, USVSN Sai and Raff, Edward and others},
  booktitle={International Conference on Machine Learning},
  pages={2397--2430},
  year={2023},
  organization={PMLR}
}

@inproceedings{boguraev2025causal,
    title = "Causal Interventions Reveal Shared Structure Across {E}nglish Filler{--}Gap Constructions",
    author = "Boguraev, Sasha  and
      Potts, Christopher  and
      Mahowald, Kyle",
    editor = "Christodoulopoulos, Christos  and
      Chakraborty, Tanmoy  and
      Rose, Carolyn  and
      Peng, Violet",
    booktitle = "Proceedings of the 2025 Conference on Empirical Methods in Natural Language Processing",
    month = nov,
    year = "2025",
    address = "Suzhou, China",
    publisher = "Association for Computational Linguistics",
    url = "https://aclanthology.org/2025.emnlp-main.1271/",
    doi = "10.18653/v1/2025.emnlp-main.1271",
    pages = "25021--25042",
    ISBN = "979-8-89176-332-6"
}

@inproceedings{geiger2024finding,
  title={Finding alignments between interpretable causal variables and distributed neural representations},
  author={Geiger, Atticus and Wu, Zhengxuan and Potts, Christopher and Icard, Thomas and Goodman, Noah},
  booktitle={Causal Learning and Reasoning},
  pages={160--187},
  year={2024},
  organization={PMLR}
}

@inproceedings{warstadt2023findings,
  title={Findings of the BabyLM challenge: Sample-efficient pretraining on developmentally plausible corpora},
  author={Warstadt, Alex and Mueller, Aaron and Choshen, Leshem and Wilcox, Ethan and Zhuang, Chengxu and Ciro, Juan and Mosquera, Rafael and Paranjabe, Bhargavi and Williams, Adina and Linzen, Tal and others},
  booktitle={Proceedings of the BabyLM challenge at the 27th conference on computational natural language learning},
  pages={1--34},
  year={2023}
}

@book{culicover1977formal,
  title={Formal syntax},
  author={Culicover, Peter W and Wasow, Thomas and Akmajian, Adrian},
  year={1977},
  publisher={Academic Press}
}

@inproceedings{ozaki2022well,
  title={How well do LSTM language models learn filler-gap dependencies?},
  author={Ozaki, Satoru and Yurovsky, Dan and Levin, Lori},
  booktitle={Proceedings of the Society for Computation in Linguistics 2022},
  pages={76--88},
  year={2022}
}

@article{lan2024large,
  title={Large language models and the argument from the poverty of the stimulus},
  author={Lan, Nur and Chemla, Emmanuel and Katzir, Roni},
  journal={Linguistic Inquiry},
  pages={1--28},
  year={2024},
  publisher={MIT Press journals-info@ mit. edu}
}

@inproceedings{chang2025mind,
  title={Mind the Gap: How BabyLMs Learn Filler-Gap Dependencies},
  author={Chang, Chi-Yun and Huang, Xueyang and Nasir, Humaira and Storks, Shane and Akingbade, Olawale and Dai, Huteng},
  booktitle={Proceedings of the 2025 Conference on Empirical Methods in Natural Language Processing},
  pages={15060--15076},
  year={2025}
}

@article{furrow1979mothers,
  title={Mothers' speech to children and syntactic development: Some simple relationships},
  author={Furrow, David and Nelson, Katherine and Benedict, Helen},
  journal={Journal of child language},
  volume={6},
  number={3},
  pages={423--442},
  year={1979},
  publisher={Cambridge University Press}
}

@article{roland2007frequency,
  title={Frequency of basic English grammatical structures: A corpus analysis},
  author={Roland, Douglas and Dick, Frederic and Elman, Jeffrey L},
  journal={Journal of memory and language},
  volume={57},
  number={3},
  pages={348--379},
  year={2007},
  publisher={Elsevier}
}

@inproceedings{arora2024causalgym,
    title = "{C}ausal{G}ym: Benchmarking causal interpretability methods on linguistic tasks",
    author = "Arora, Aryaman  and
      Jurafsky, Dan  and
      Potts, Christopher",
    editor = "Ku, Lun-Wei  and
      Martins, Andre  and
      Srikumar, Vivek",
    booktitle = "Proceedings of the 62nd Annual Meeting of the Association for Computational Linguistics (Volume 1: Long Papers)",
    month = aug,
    year = "2024",
    address = "Bangkok, Thailand",
    publisher = "Association for Computational Linguistics",
    url = "https://aclanthology.org/2024.acl-long.785/",
    doi = "10.18653/v1/2024.acl-long.785",
    pages = "14638--14663"
}

@article{wilcox2024using,
  title={Using computational models to test syntactic learnability},
  author={Wilcox, Ethan Gotlieb and Futrell, Richard and Levy, Roger},
  journal={Linguistic Inquiry},
  volume={55},
  number={4},
  pages={805--848},
  year={2024},
  publisher={MIT Press journals-info@ mit. edu}
}

@article{schutze_challenges_2015,
	title = {Challenges for a theory of islands: A broader perspective on Ambridge, Pine, and Lieven},
	volume = {91},
	issn = {1535-0665},
	url = {https://muse.jhu.edu/pub/24/article/583509},
	shorttitle = {Challenges for a theory of islands},
	pages = {31--39},
	number = {2},
	journal = {Language},
	author = {Schütze, Carson T. and Sprouse, Jon and Caponigro, Ivano},
	urldate = {2024-10-03},
	date = {2015},
        year = {2015},
}

@inproceedings{kobzeva_neural_2023,
	address = {Amherst, MA},
	title = {Neural {Networks} {Can} {Learn} {Patterns} of {Island}-insensitivity in {Norwegian}},
	url = {https://aclanthology.org/2023.scil-1.15},
	urldate = {2024-01-19},
	booktitle = {Proceedings of the {Society} for {Computation} in {Linguistics} 2023},
	publisher = {Association for Computational Linguistics},
	author = {Kobzeva, Anastasia and Arehalli, Suhas and Linzen, Tal and Kush, Dave},
	editor = {Hunter, Tim and Prickett, Brandon},
	month = jun,
	year = {2023},
	pages = {175--185},
}

@inproceedings{suijkerbuijk2023learnability,
  title={The learnability of the wh-island constraint in dutch by a long short-term memory network},
  author={Suijkerbuijk, Michelle and de Swart, Peter and Frank, Stefan L},
  booktitle={Proceedings of the Society for Computation in Linguistics 2023},
  pages={321--331},
  year={2023}
}

@article{chomsky_wh-movement_1977,
	title = {On {Wh}-{Movement}},
	url = {https://cir.nii.ac.jp/crid/1574231874477886720},
	urldate = {2024-01-19},
	journal = {Formal Syntax},
	author = {Chomsky, Noam},
	year = {1977},
	note = {Publisher: Academic Press},
	pages = {71--132},
}

@article{huang2024large,
  title={Large-scale benchmark yields no evidence that language model surprisal explains syntactic disambiguation difficulty},
  author={Huang, Kuan-Jung and Arehalli, Suhas and Kugemoto, Mari and Muxica, Christian and Prasad, Grusha and Dillon, Brian and Linzen, Tal},
  journal={Journal of Memory and Language},
  volume={137},
  pages={104510},
  year={2024},
  publisher={Elsevier}
}

@inproceedings{futrell2019neural,
  title={Neural Language Models as Psycholinguistic Subjects: Representations of Syntactic State},
  author={Futrell, Richard and Wilcox, Ethan and Morita, Takashi and Qian, Peng and Ballesteros, Miguel and Levy, Roger},
  booktitle={Proceedings of NAACL-HLT},
  pages={32--42},
  year={2019}
}

@article{levy2008expectation,
  title={Expectation-based syntactic comprehension},
  author={Levy, Roger},
  journal={Cognition},
  volume={106},
  number={3},
  pages={1126--1177},
  year={2008},
  publisher={Elsevier}
}

@article{meylan2023adults,
  title={How adults understand what young children say},
  author={Meylan, Stephan C and Foushee, Ruthe and Wong, Nicole H and Bergelson, Elika and Levy, Roger P},
  journal={Nature human behaviour},
  volume={7},
  number={12},
  pages={2111--2125},
  year={2023},
  publisher={Nature Publishing Group UK London}
}

@article{vong2024grounded,
  title={Grounded language acquisition through the eyes and ears of a single child},
  author={Vong, Wai Keen and Wang, Wentao and Orhan, A Emin and Lake, Brenden M},
  journal={Science},
  volume={383},
  number={6682},
  pages={504--511},
  year={2024},
  publisher={American Association for the Advancement of Science}
}

@article{traxler1996plausibility,
  title={Plausibility and the processing of unbounded dependencies: An eye-tracking study},
  author={Traxler, Matthew J. and Pickering, Martin J.},
  journal={Journal of Memory and Language},
  volume={35},
  pages={454--475},
  year={1996}
}

@article{crain_rules_1985,
	title = {Rules and Constraints in Sentence Processing},
	volume = {15},
	number = {1},
	journal = {North East Linguistics Society},
	author = {Crain, Stephen and Fodor, Janet Dean},
	date = {1985},
}

@article{sprouse2016experimental,
  title={Experimental syntax and the variation of island effects in English and Italian},
  author={Sprouse, Jon and Caponigro, Ivano and Greco, Ciro and Cecchetto, Carlo},
  journal={Natural Language \& Linguistic Theory},
  volume={34},
  number={1},
  pages={307--344},
  year={2016},
  publisher={Springer}
}

@article{friedmann2009relativized,
  title={Relativized relatives: Types of intervention in the acquisition of A-bar dependencies},
  author={Friedmann, Naama and Belletti, Adriana and Rizzi, Luigi},
  journal={Lingua},
  volume={119},
  number={1},
  pages={67--88},
  year={2009},
  publisher={Elsevier}
}

@article{kush2021sentence,
  title={Sentence processing and syntactic theory},
  author={Kush, Dave and Dillon, Brian},
  journal={A companion to Chomsky},
  pages={305--324},
  year={2021},
  publisher={Wiley Online Library}
}

@inproceedings{bosch2025another,
  title={On another topic, how do acquisition orders vary? the left-periphery and topicalization in bilingual and monolingual acquisition},
  author={Bosch, N{\'u}ria and Biberauer, Theresa},
  booktitle={Proceedings of the 49th Boston University Conference on Language Development (BUCLD)},
  pages={129--144},
  year={2025},
  organization={Cascadilla Proceedings Project Somerville, MA}
}

@article{de1995relative,
  title={Relative clauses are barriers to wh-movement for young children},
  author={De Villiers, Jill and Roeper, Thomas},
  journal={Journal of child Language},
  volume={22},
  number={2},
  pages={389--404},
  year={1995},
  publisher={Cambridge University Press}
}

@article{yang2026unified,
  title={A Unified Assessment of the Poverty of the Stimulus Argument for Neural Language Models},
  author={Yang, Xiulin and Bisazza, Arianna and Schneider, Nathan and Wilcox, Ethan Gotlieb},
  journal={arXiv preprint arXiv:2602.09992},
  year={2026}
}

@article{zhou2026exactly,
  title={What Exactly do Children Receive in Language Acquisition? A Case Study on CHILDES with Automated Detection of Filler-Gap Dependencies},
  author={Zhou, Zhenghao Herbert and Dai, William and Viswanathan, Maya and Charlow, Simon and McCoy, R Thomas and Frank, Robert},
  journal={arXiv preprint arXiv:2603.02082},
  year={2026}
}

@article{perkins2025mind,
  title={Mind the gap: Learning the surface forms of movement dependencies},
  author={Perkins, Laurel and Feldman, Naomi H and Lidz, Jeffrey},
  journal={Language},
  pages={1--42},
  year={2025},
  publisher={Cambridge University Press}
}

@article{portelance2025reframing,
  title={Reframing linguistic bootstrapping as joint inference using visually-grounded grammar induction models},
  author={Portelance, Eva and Reddy, Siva and O’Donnell, Timothy J},
  journal={Journal of Memory and Language},
  volume={145},
  pages={104672},
  year={2025},
  publisher={Elsevier}
}

@inproceedings{bhattacharya-van-schijndel-2020-filler,
    title = "Filler-gaps that neural networks fail to generalize",
    author = "Bhattacharya, Debasmita  and
      van Schijndel, Marten",
    editor = "Fern{\'a}ndez, Raquel  and
      Linzen, Tal",
    booktitle = "Proceedings of the 24th Conference on Computational Natural Language Learning",
    month = nov,
    year = "2020",
    address = "Online",
    publisher = "Association for Computational Linguistics",
    url = "https://aclanthology.org/2020.conll-1.39/",
    doi = "10.18653/v1/2020.conll-1.39",
    pages = "486--495"
}

@inproceedings{prasad2019using,
  title={Using Priming to Uncover the Organization of Syntactic Representations in Neural Language Models},
  author={Prasad, Grusha and van Schijndel, Marten and Linzen, Tal},
  booktitle={Proceedings of the 23rd Conference on Computational Natural Language Learning (CoNLL)},
  pages={66--76},
  year={2019}
}

@inproceedings{hao2023verb,
  title={Verb Conjugation in Transformers Is Determined by Linear Encodings of Subject Number},
  author={Hao, Sophie and Linzen, Tal},
  booktitle={Findings of the Association for Computational Linguistics: EMNLP 2023},
  pages={4531--4539},
  year={2023}
}

@inproceedings{lasri2022probing,
  title={Probing for the Usage of Grammatical Number},
  author={Lasri, Karim and Pimentel, Tiago and Lenci, Alessandro and Poibeau, Thierry and Cotterell, Ryan},
  booktitle={Proceedings of the 60th Annual Meeting of the Association for Computational Linguistics (Volume 1: Long Papers)},
  pages={8818--8831},
  year={2022}
}

@inproceedings{wanginterpretability,
  title={Interpretability in the Wild: a Circuit for Indirect Object Identification in GPT-2 Small},
  year={2021},
  author={Wang, Kevin Ro and Variengien, Alexandre and Conmy, Arthur and Shlegeris, Buck and Steinhardt, Jacob},
  booktitle={The Eleventh International Conference on Learning Representations}
}

@misc{kryvosheieva2025different,
      title={Different types of syntactic agreement recruit the same units within large language models},
      author={Daria Kryvosheieva and Andrea de Varda and Evelina Fedorenko and Greta Tuckute},
      year={2025},
      eprint={2512.03676},
      archivePrefix={arXiv},
      primaryClass={cs.CL},
      url={https://arxiv.org/abs/2512.03676},
}

@inproceedings{gauthier2020syntaxgym,
  title={SyntaxGym: An online platform for targeted evaluation of language models},
  author={Gauthier, Jon and Hu, Jennifer and Wilcox, Ethan and Qian, Peng and Levy, Roger},
  booktitle={Proceedings of the 58th Annual Meeting of the Association for Computational Linguistics: System Demonstrations},
  pages={70--76},
  year={2020}
}

@article{gao2020pile,
  title={The {P}ile: An 800GB Dataset of Diverse Text for Language Modeling},
  author={Gao, Leo and Biderman, Stella and Black, Sid and Golding, Laurence and Hoppe, Travis and Foster, Charles and Phang, Jason and He, Horace and Thite, Anish and Nabeshima, Noa and Presser, Shawn and Leahy, Connor},
  journal={arXiv preprint arXiv:2101.00027},
  year={2020}
}

@article{perkins2021eighteen,
  title={Eighteen-month-old infants represent nonlocal syntactic dependencies},
  author={Perkins, Laurel and Lidz, Jeffrey},
  journal={Proceedings of the National Academy of Sciences},
  volume={118},
  number={41},
  pages={e2026469118},
  year={2021},
  publisher={National Academy of Sciences}
}

@article{atkinson2018developing,
  title={Developing incrementality in filler-gap dependency processing},
  author={Atkinson, Emily and Wagers, Matthew W and Lidz, Jeffrey and Phillips, Colin and Omaki, Akira},
  journal={Cognition},
  volume={179},
  pages={132--149},
  year={2018},
  publisher={Elsevier}
}

@article{gagliardi2016discontinuous,
  title={Discontinuous development in the acquisition of filler-gap dependencies: Evidence from 15-and 20-month-olds},
  author={Gagliardi, Annie and Mease, Tara M and Lidz, Jeffrey},
  journal={Language Acquisition},
  volume={23},
  number={3},
  pages={234--260},
  year={2016},
  publisher={Taylor \& Francis}
}

@article{radford2019language,
  title={Language models are unsupervised multitask learners},
  author={Radford, Alec and Wu, Jeffrey and Child, Rewon and Luan, David and Amodei, Dario and Sutskever, Ilya and others},
  journal={OpenAI blog},
  volume={1},
  number={8},
  pages={9},
  year={2019}
}

@inproceedings{giulianelli_proper_2024,
    address = {Miami, Florida, USA},
    title = {On the {Proper} {Treatment} of {Tokenization} in {Psycholinguistics}},
    url = {https://aclanthology.org/2024.emnlp-main.1032/},
    doi = {10.18653/v1/2024.emnlp-main.1032},
    urldate = {2025-04-30},
    booktitle = {Proceedings of the 2024 {Conference} on {Empirical} {Methods} in {Natural} {Language} {Processing}},
    publisher = {Association for Computational Linguistics},
    author = {Giulianelli, Mario and Malagutti, Luca and Gastaldi, Juan Luis and DuSell, Brian and Vieira, Tim and Cotterell, Ryan},
    editor = {Al-Onaizan, Yaser and Bansal, Mohit and Chen, Yun-Nung},
    month = nov,
    year = {2024},
    pages = {18556--18572},
}

@inproceedings{nair2023words,
  title={Words, Subwords, and Morphemes: What Really Matters in the Surprisal-Reading Time Relationship?},
  author={Nair, Sathvik and Resnik, Philip},
  booktitle={Findings of the Association for Computational Linguistics: EMNLP 2023},
  pages={11251--11260},
  year={2023}
}

@inproceedings{oh2025impact,
  title={The impact of token granularity on the predictive power of language model surprisal},
  author={Oh, Byung-Doh and Schuler, William},
  booktitle={Proceedings of the 63rd Annual Meeting of the Association for Computational Linguistics (Volume 1: Long Papers)},
  pages={4150--4162},
  year={2025}
}

@misc{charpentier2025babylmturns3papers,
      title={BabyLM Turns 3: Call for papers for the 2025 BabyLM workshop}, 
      author={Lucas Charpentier and Leshem Choshen and Ryan Cotterell and Mustafa Omer Gul and Michael Hu and Jaap Jumelet and Tal Linzen and Jing Liu and Aaron Mueller and Candace Ross and Raj Sanjay Shah and Alex Warstadt and Ethan Wilcox and Adina Williams},
      year={2025},
      eprint={2502.10645},
      archivePrefix={arXiv},
      primaryClass={cs.CL},
      url={https://arxiv.org/abs/2502.10645}, 
}

@article{gilkerson2017mapping,
  title={Mapping the early language environment using all-day recordings and automated analysis},
  author={Gilkerson, Jill and Richards, Jeffrey A and Warren, Steven F and Montgomery, Judith K and Greenwood, Charles R and Kimbrough Oller, D and Hansen, John HL and Paul, Terrance D},
  journal={American journal of speech-language pathology},
  volume={26},
  number={2},
  pages={248--265},
  year={2017},
  publisher={American Speech-Language-Hearing Association}
}

@article{perkins2022power,
  title={The power of ignoring: Filtering input for argument structure acquisition},
  author={Perkins, Laurel and Feldman, Naomi H and Lidz, Jeffrey},
  journal={Cognitive Science},
  volume={46},
  number={1},
  pages={e13080},
  year={2022},
  publisher={Wiley Online Library}
}

@article{portelance2024roles,
  title={The roles of neural networks in language acquisition},
  author={Portelance, Eva and Jasbi, Masoud},
  journal={Language and Linguistics Compass},
  volume={18},
  number={6},
  pages={e70001},
  year={2024},
  publisher={Wiley Online Library}
}

@inproceedings{murty2023pushdown,
  title={Pushdown Layers: Encoding Recursive Structure in Transformer Language Models},
  author={Murty, Shikhar and Sharma, Pratyusha and Andreas, Jacob and Manning, Christopher D},
  booktitle={Proceedings of the 2023 Conference on Empirical Methods in Natural Language Processing},
  pages={3233--3247},
  year={2023}
}

@article{pearl2023computational,
  title={Computational cognitive modeling for syntactic acquisition: Approaches that integrate information from multiple places},
  author={Pearl, Lisa},
  journal={Journal of Child Language},
  volume={50},
  number={6},
  pages={1353--1373},
  year={2023},
  publisher={Cambridge University Press}
}

@article{piantadosi2023modern,
  title={Modern language models refute Chomsky’s approach to language},
  author={Piantadosi, Steven T},
  journal={From fieldwork to linguistic theory: A tribute to Dan Everett},
  volume={15},
  pages={353--414},
  year={2023},
  publisher={Language Science Press Berlin, Germany}
}

@article{futrell2025linguistics,
  title={How linguistics learned to stop worrying and love the language models},
  author={Futrell, Richard and Mahowald, Kyle},
  journal={arXiv preprint arXiv:2501.17047},
  year={2025}
}

@article{warstadt2020blimp,
  title={BLiMP: The benchmark of linguistic minimal pairs for English},
  author={Warstadt, Alex and Parrish, Alicia and Liu, Haokun and Mohananey, Anhad and Peng, Wei and Wang, Sheng-Fu and Bowman, Samuel R},
  journal={Transactions of the Association for Computational Linguistics},
  volume={8},
  pages={377--392},
  year={2020},
  publisher={MIT Press One Rogers Street, Cambridge, MA 02142-1209, USA journals-info~…}
}

@inproceedings{wu-etal-2024-pyvene,
    title = "pyvene: A Library for Understanding and Improving {P}y{T}orch Models via Interventions",
    author = "Wu, Zhengxuan and Geiger, Atticus and Arora, Aryaman and Huang, Jing and Wang, Zheng and Goodman, Noah and Manning, Christopher and Potts, Christopher",
    editor = "Chang, Kai-Wei and Lee, Annie and Rajani, Nazneen",
    booktitle = "Proceedings of the 2024 Conference of the North American Chapter of the Association for Computational Linguistics: Human Language Technologies (Volume 3: System Demonstrations)",
    month = jun,
    year = "2024",
    address = "Mexico City, Mexico",
    publisher = "Association for Computational Linguistics",
    url = "https://aclanthology.org/2024.naacl-demo.16",
    pages = "158--165",
}

@misc{Lenth2017,
  title = {emmeans: Estimated Marginal Means,  aka Least-Squares Means},
  url = {http://dx.doi.org/10.32614/CRAN.package.emmeans},
  DOI = {10.32614/cran.package.emmeans},
  journal = {CRAN: Contributed Packages},
  publisher = {The R Foundation},
  author = {Lenth,  Russell V. and Piaskowski,  Julia},
  year = {2017},
  month = oct 
}

@incollection{gazdar_phrase_1982,
	address = {Dordrecht},
	series = {Synthese {Language} {Library}},
	title = {Phrase {Structure} {Grammar}},
	isbn = {978-94-009-7707-5},
	url = {https://doi.org/10.1007/978-94-009-7707-5_5},
	language = {en},
	urldate = {2024-01-19},
	booktitle = {The {Nature} of {Syntactic} {Representation}},
	publisher = {Springer Netherlands},
	author = {Gazdar, Gerald},
	editor = {Jacobson, Pauline and Pullum, Geoffrey K.},
	year = {1982},
	doi = {10.1007/978-94-009-7707-5_5},
	pages = {131--186},
}

@incollection{kaplan_lexical-functional_1982,
	title = {Lexical-{Functional} {Grammar}: {A} {Formal} {System} for {Grammatical} {Representation}},
	shorttitle = {Lexical-{Functional} {Grammar}},
	booktitle = {The {Mental} {Representation} of {Grammatical} {Relations}},
	publisher = {MIT Press},
	author = {Kaplan, Ronald and Bresnan, Joan},
	editor = {Bresnan, Joan},
	month = jan,
	year = {1982},
	pages = {173--281}}

@book{postal_three_1999,
	title = {Three {Investigations} of {Extraction}},
	isbn = {978-0-262-28181-2},
	url = {https://direct.mit.edu/books/book/4699/Three-Investigations-of-Extraction},
	language = {en},
	urldate = {2024-01-19},
	publisher = {The MIT Press},
	author = {Postal, Paul M.},
	month = jan,
	year = {1999},
	doi = {10.7551/mitpress/6820.001.0001},
}

@article{linzen2021syntactic,
  title={Syntactic structure from deep learning},
  author={Linzen, Tal and Baroni, Marco},
  journal={Annual Review of Linguistics},
  volume={7},
  pages={195--212},
  year={2021},
  publisher={Annual Reviews}
}

@article{mueller2025quest,
  title={The Quest for the Right Mediator: Surveying Mechanistic Interpretability for NLP Through the Lens of Causal Mediation Analysis},
  author={Mueller, Aaron and Brinkmann, Jannik and Li, Millicent and Marks, Samuel and Pal, Koyena and Prakash, Nikhil and Rager, Can and Sankaranarayanan, Aruna and Sharma, Arnab Sen and Sun, Jiuding and others},
  journal={Computational Linguistics},
  pages={1--48},
  year={2025},
  publisher={MIT Press 255 Main Street, 9th Floor, Cambridge, Massachusetts 02142, USA~…}
}

@article{wilcox2025bigger,
  title={Bigger is not always better: The importance of human-scale language modeling for psycholinguistics},
  author={Wilcox, Ethan Gotlieb and Hu, Michael Y and Mueller, Aaron and Warstadt, Alex and Choshen, Leshem and Zhuang, Chengxu and Williams, Adina and Cotterell, Ryan and Linzen, Tal},
  journal={Journal of Memory and Language},
  volume={144},
  pages={104650},
  year={2025},
  publisher={Elsevier}
}

@article{yang2004universal,
  title={Universal Grammar, statistics or both?},
  author={Yang, Charles D},
  journal={Trends in cognitive sciences},
  volume={8},
  number={10},
  pages={451--456},
  year={2004},
  publisher={Elsevier}
}

@inproceedings{marvin2018targeted,
  title={Targeted Syntactic Evaluation of Language Models},
  author={Marvin, Rebecca and Linzen, Tal},
  booktitle={Proceedings of the 2018 Conference on Empirical Methods in Natural Language Processing},
  pages={1192--1202},
  year={2018}
}

@article{misra2024generating,
  title={Generating novel experimental hypotheses from language models: A case study on cross-dative generalization},
  author={Misra, Kanishka and Kim, Najoung},
  journal={arXiv preprint arXiv:2408.05086},
  year={2024}
}

@article{traxler2014syntactic,
  title={Syntactic priming during sentence comprehension: Evidence for the lexical boost.},
  author={Traxler, Matthew J and Tooley, Kristen M and Pickering, Martin J},
  journal={Journal of Experimental Psychology: Learning, Memory, and Cognition},
  volume={40},
  number={4},
  pages={905},
  year={2014},
  publisher={American Psychological Association}
}

@article{wu2023interpretability,
  title={Interpretability at scale: Identifying causal mechanisms in alpaca},
  author={Wu, Zhengxuan and Geiger, Atticus and Icard, Thomas and Potts, Christopher and Goodman, Noah},
  journal={Advances in neural information processing systems},
  volume={36},
  pages={78205--78226},
  year={2023}
}
\clearpage

\appendix
\section{Appendix} \label{section:appendix}
\subsection{Hyperparameter Selection}
\label{sec:hparam}

Early experiments found that the default hyperparameters reported in the ~\citet{boguraev2025causal} codebase (batch size 25 $\times$ 16 steps) resulted in undertrained DAS vectors. 
This is possible because the Pythia 1.4B model has far more parameters and was trained on far more data compared to BabyLM-100M.

\begin{figure*}
\centering
\includegraphics[width=\textwidth]{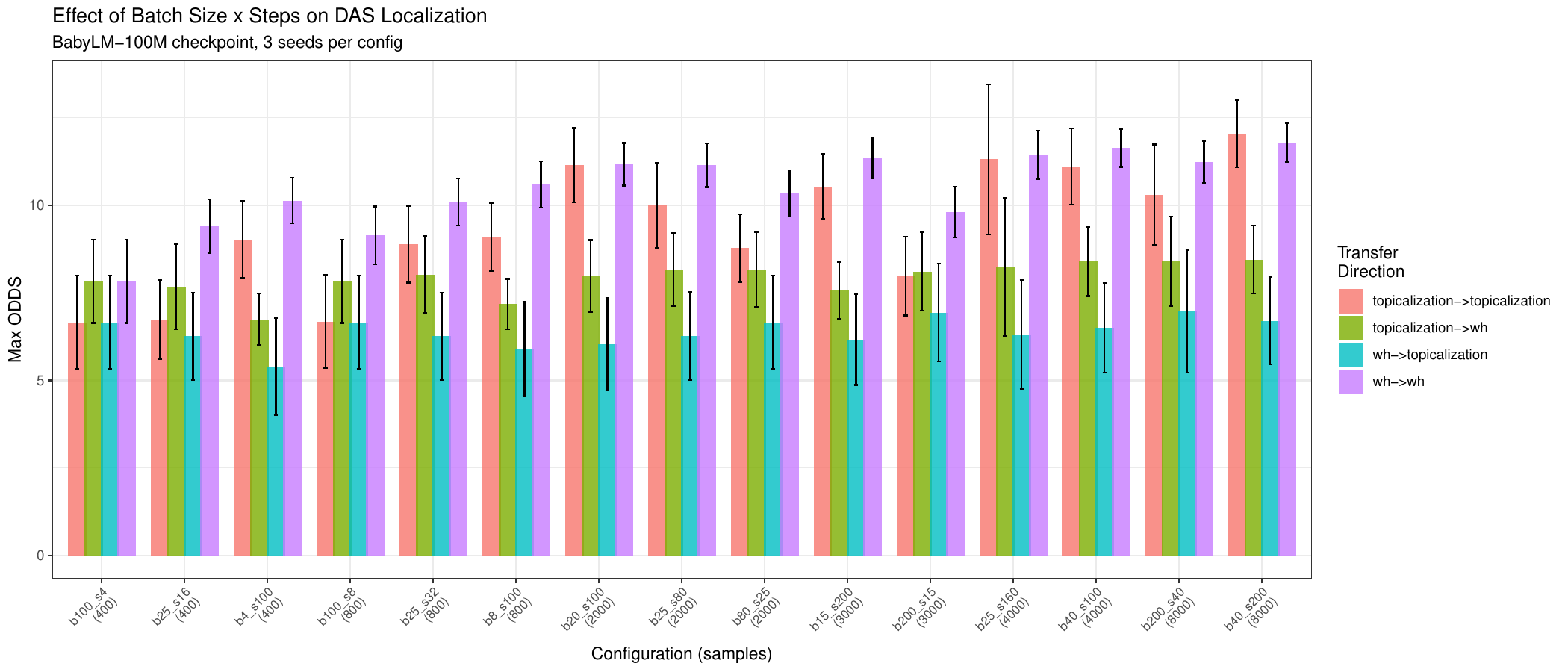}
\caption{Hyperparameter sweep for DAS training. \maxodds increases with training samples and stabilizes around 2000 samples (batch size 25 $\times$ 80 steps).}
\label{fig:hparam_main}
\end{figure*}

\begin{figure*}
\centering
\includegraphics[width=\textwidth]{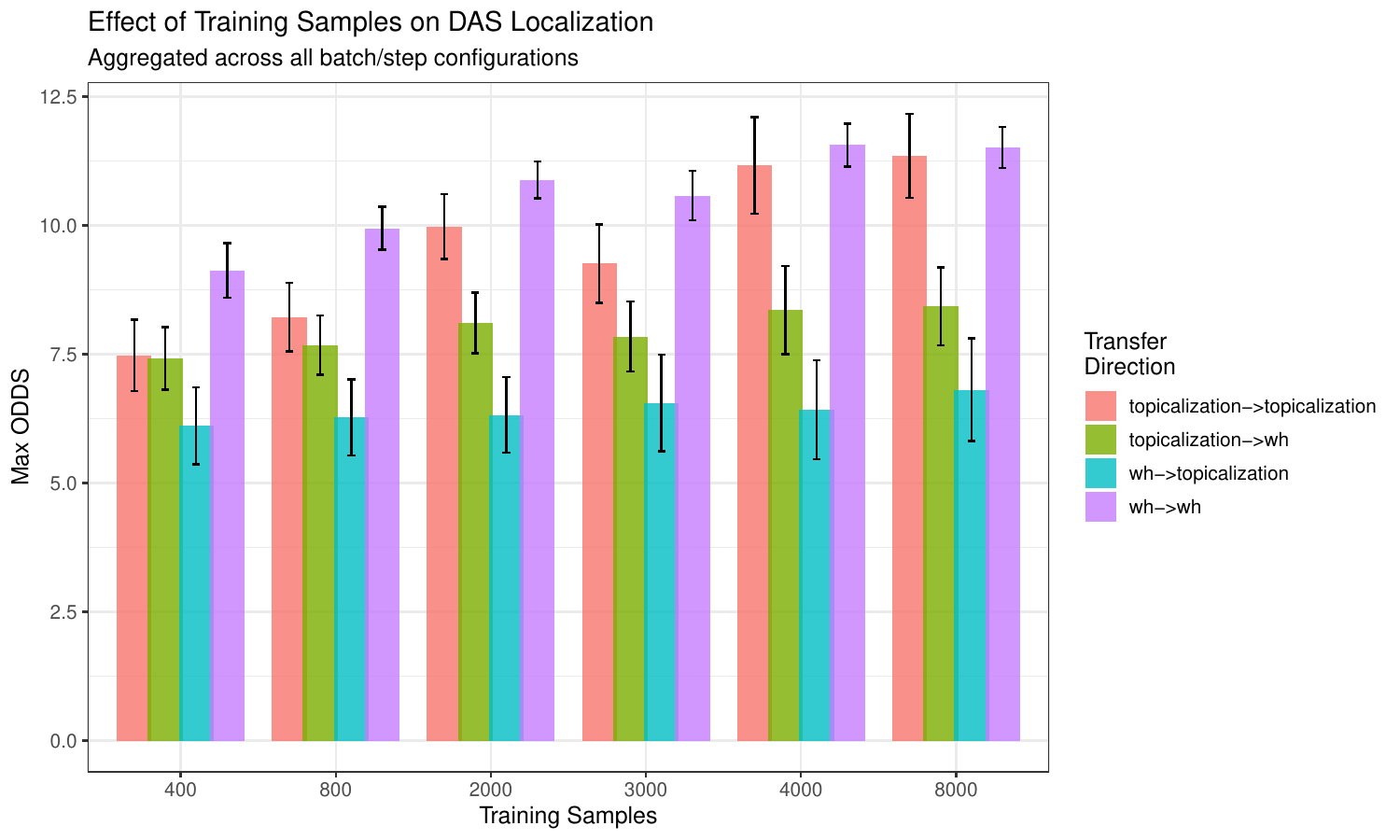}
\caption{\maxodds as a function of total training samples, collapsed across batch sizes. Performance plateaus around 2000--2500 samples.}
\label{fig:hparam_samples}
\end{figure*}

We conducted a hyperparameter sweep to determine optimal DAS training parameters. Figure~\ref{fig:hparam_main} and ~\ref{fig:hparam_samples} show \maxodds across different batch sizes (8, 16, 25, 32) and training steps (40, 60, 80, 100, 120) for the Wh$\rightarrow$Wh within-construction condition at the 100M checkpoint. Based on these results, we selected a batch size of 25 with 80 training steps (2000 total samples) for all experiments.

The learning rate was fixed at the default value of  5$\times$10$^{-3}$ used in \citet{arora2024causalgym}.

\subsection{Animacy Figures}
\label{sec:animacyfigs}
\begin{figure*}
\centering
\includegraphics[width=\textwidth]{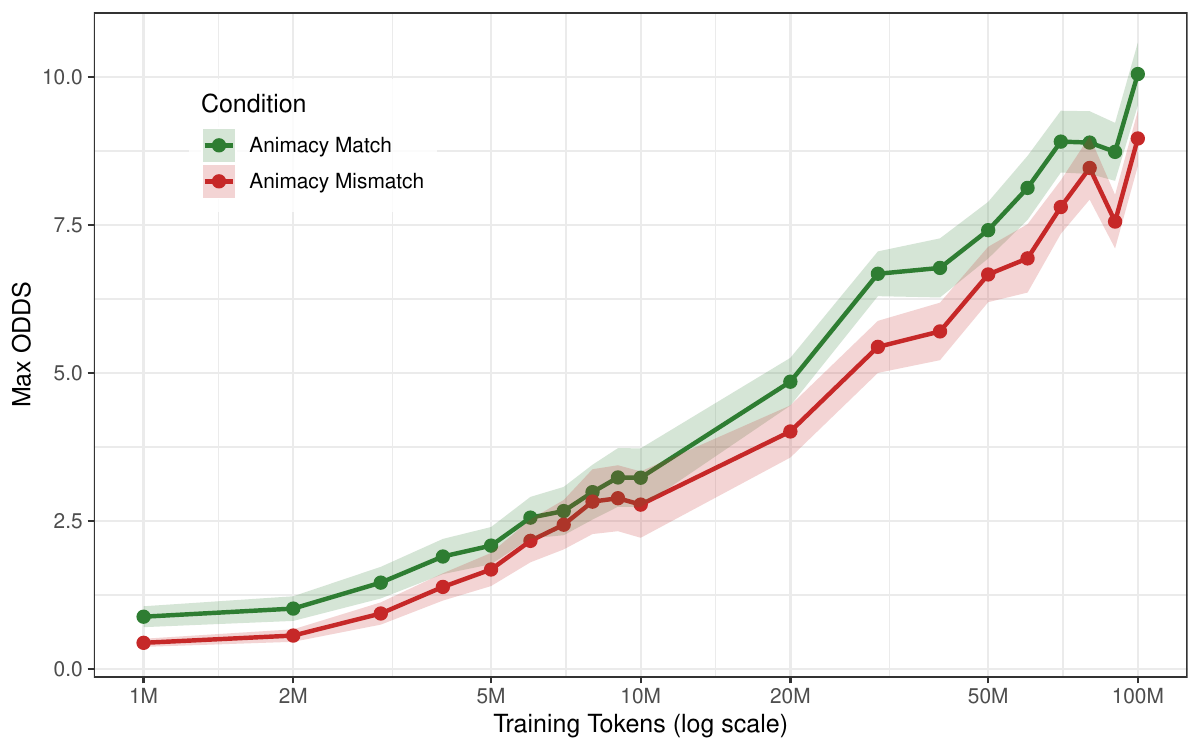} 
\caption{
Developmental trajectory of lexical boost across training. Error bands show $\pm$1 SE across a minimum of 2 seeds.}
\label{fig:animacy_trajectory}
\end{figure*}

Supplementing the statistical tests for animacy effects in \ref{sec:animacy_effects}, we plot the increase in \maxodds across training split by lexical matching conditions. Figure~\ref{fig:animacy_trajectory} compares the causal performance when the intervention source and target base sentences share the same animacy status (Animate $\to$ Animate) versus differing animacy status (Animate $\to$ Inanimate). Results show a consistent gap between the two conditions, suggesting the learned representation may retain sensitivity to lexical features, such as animacy, through the pretraining process. 

To separate animacy effects from construction-specific variance, the reported metrics for both Figure~\ref{fig:animacy_trajectory} and statistical testing are averaged across wh-question and topicalization constructions.  

\subsection{Beyond Developmental Constraints}
\label{sec:extended}

We also present our results for the full 1 billion token training trajectory (100M--1000M tokens)\footnote{The model received 1000M tokens because it was trained on the 100M token dataset for 10 epochs.} based on additional checkpoints released for the BabyLM model to better understand how filler-gap mechanisms continue to develop beyond developmentally plausible input levels. 
We see cross-construction generalizations plateau after 100M tokens, and performance begins to overlap for both sets of results. LMs could require more input to learn a representation specific to topicalization, since it rarely shows up in children's input.
These results confirm our overall claims localization within examples of one construction is stronger than transfer across constructions, while showing inconclusive evidence for our third hypothesis regarding interactions between construction frequency and transfer.

\begin{figure*}
\centering
\includegraphics[width=\textwidth]{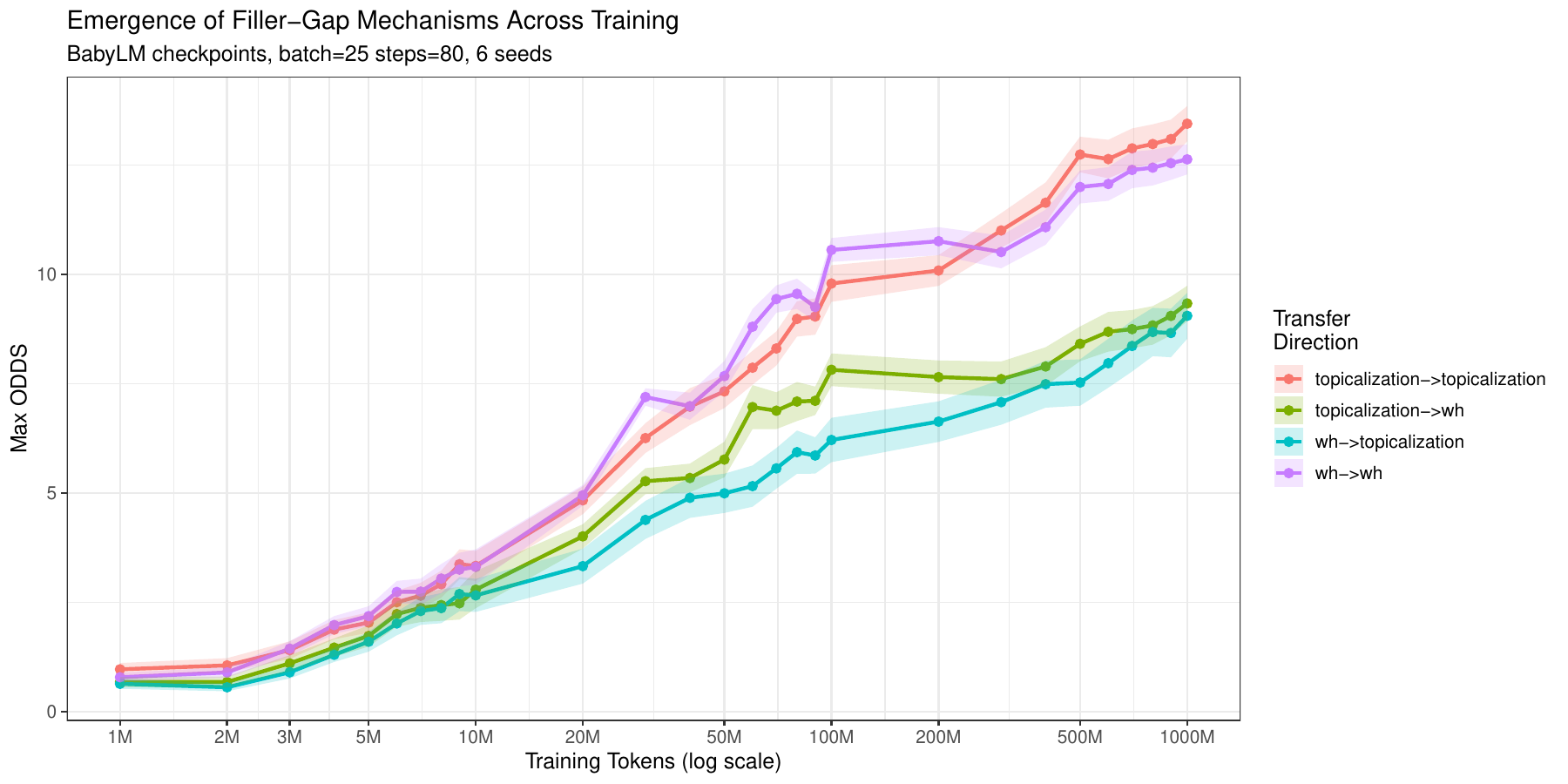}
\caption{Full developmental trajectory from 1M to 1000M tokens. Filler-gap mechanisms continue to improve but begin to plateau around 500M--700M tokens.}
\label{fig:long_emergence}
\end{figure*}

\begin{figure*}
\centering
\includegraphics[width=\textwidth]{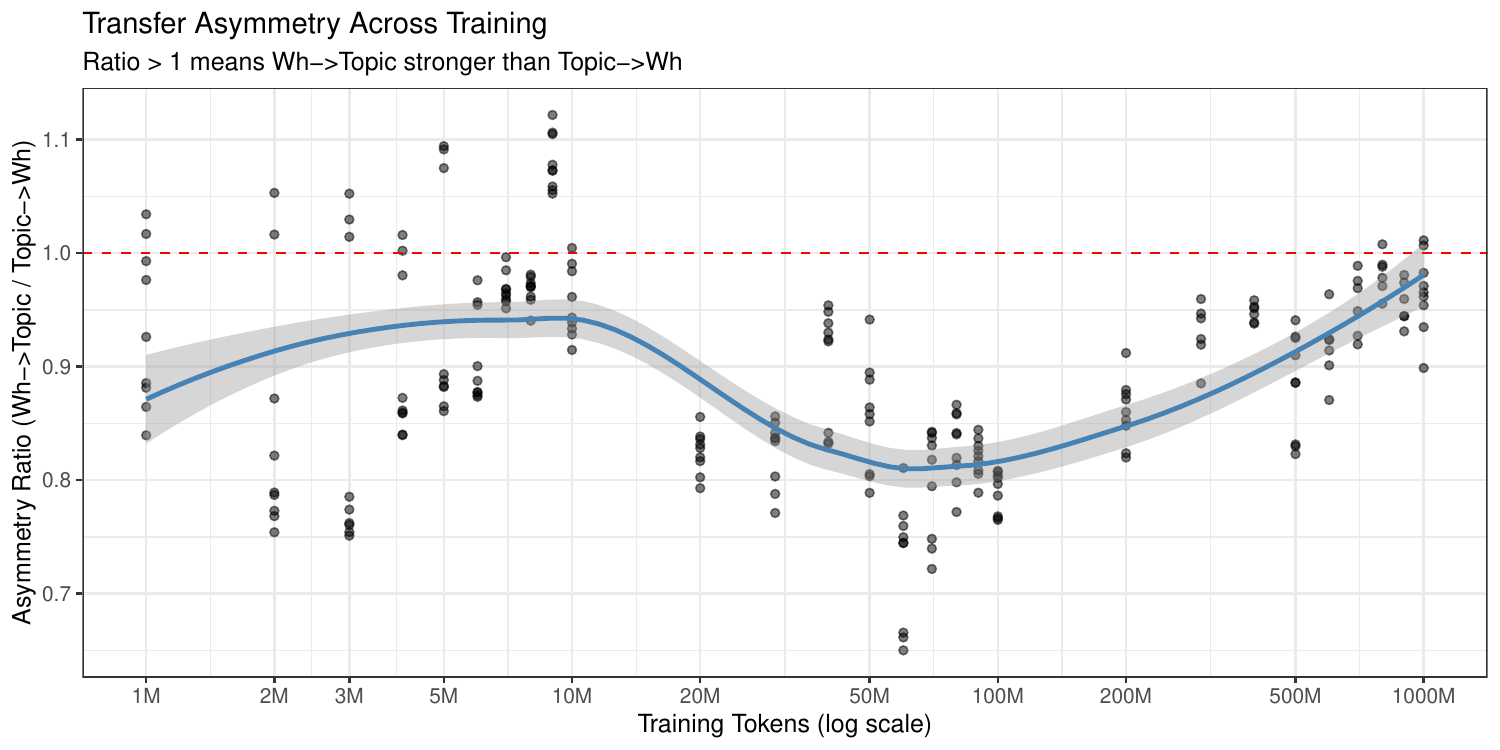}
\caption{Transfer asymmetry across full training range. The asymmetry (Topic$\rightarrow$Wh $>$ Wh$\rightarrow$Topic) persists and slightly increases at later checkpoints.}
\label{fig:long_asymmetry}
\end{figure*}

\end{document}